\documentclass[10pt,twocolumn,letterpaper]{article}

\usepackage{cvpr}
\usepackage{siunitx}
\usepackage{times}
\usepackage{epsfig}
\usepackage{graphicx}
\usepackage{amsmath}
\usepackage{amssymb}
\usepackage{booktabs}

\usepackage[pagebackref=true,breaklinks=true,letterpaper=true,colorlinks,bookmarks=false]{hyperref}

\cvprfinalcopy 

\def\cvprPaperID{1358} 

\begin{document}

\title{Dynamic Edge-Conditioned Filters in Convolutional Neural Networks on Graphs}
\author{Martin~Simonovsky\\
Universit\'{e} Paris Est, \'{E}cole des Ponts ParisTech\\
{\tt\small martin.simonovsky@enpc.fr}
\and
Nikos~Komodakis\\
Universit\'{e} Paris Est, \'{E}cole des Ponts ParisTech\\
{\tt\small nikos.komodakis@enpc.fr}
}

\maketitle

\begin{abstract}
A number of problems can be formulated as prediction on graph-structured data. In this work, we generalize the convolution operator from regular grids to arbitrary graphs while avoiding the spectral domain, which allows us to handle graphs of varying size and connectivity. To move beyond a simple diffusion, filter weights are conditioned on the specific edge labels in the neighborhood of a vertex. Together with the proper choice of graph coarsening, we explore constructing deep neural networks for graph classification. In particular, we demonstrate the generality of our formulation in point cloud classification, where we set the new state of the art, and on a graph classification dataset, where we outperform other deep learning approaches. The source code is available at \url{https://github.com/mys007/ecc}.
\end{abstract}

\def\eg{\textit{e.g.}~}
\def\Eg{\textit{E.g.}~}
\def\etal{\textit{et al.}\xspace}
\def\ie{\textit{i.e.}~}
\def\cf{\textit{c.f.}~}
\def\wrt{{w.r.t.}~}

\section{Introduction}

Convolutional Neural Networks (CNNs) have gained massive popularity in tasks where the underlying data representation has a grid structure, such as in speech processing and natural language understanding (1D, temporal convolutions), in image classification and segmentation (2D, spatial convolutions), or in video parsing (3D, volumetric convolutions)~\cite{lecun2015deep}. 

On the other hand, in many other tasks the data naturally lie on irregular or generally non-Euclidean domains, which can be structured as graphs in many cases. These include problems in 3D modeling, computational chemistry and biology, geospatial analysis, social networks, or natural language semantics and knowledge bases, to name a few. Assuming that the locality, stationarity, and composionality principles of representation hold to at least some level in the data, it is meaningful to consider a hierarchical CNN-like architecture for processing it.

However, a generalization of CNNs from grids to general graphs is not straightforward and has recently become a topic of increased interest. We identify that the current formulations of graph convolution do not exploit edge labels, which results in an overly homogeneous view of local graph neighborhoods, with an effect similar to enforcing rotational invariance of filters in regular convolutions on images. Hence, in this work we propose a convolution operation which can make use of this information channel and show that it leads to an improved graph classification performance.

This novel formulation also opens up a broader range of applications; we concentrate here on point clouds specifically. Point clouds have been mostly ignored by deep learning so far, their voxelization being the only trend to the best of our knowledge \cite{voxnet,pclabeling16}. To offer a competitive alternative with a different set of advantages and disadvantages, we construct graphs in Euclidean space from point clouds in this work and demonstrate state of the art performance on Sydney dataset of LiDAR scans~\cite{trianglesvm}.

Our contributions are as follows:

\begin{itemize}
\item We formulate a convolution-like operation on graph signals performed in the spatial domain where filter weights are conditioned on edge labels (discrete or continuous) and dynamically generated for each specific input sample. Our networks work on graphs with arbitrary varying structure throughout a dataset.
\item We are the first to apply graph convolutions to point cloud classification. Our method outperforms volumetric approaches and attains the new state of the art performance on Sydney dataset, with the benefit of preserving sparsity and presumably fine details.
\item We reach a competitive level of performance on graph classification benchmark NCI1~\cite{nci1db}, outperforming other approaches based on deep learning there.
\end{itemize}

\begin{figure*}[ht]
\centering
\includegraphics[width=0.65\linewidth]{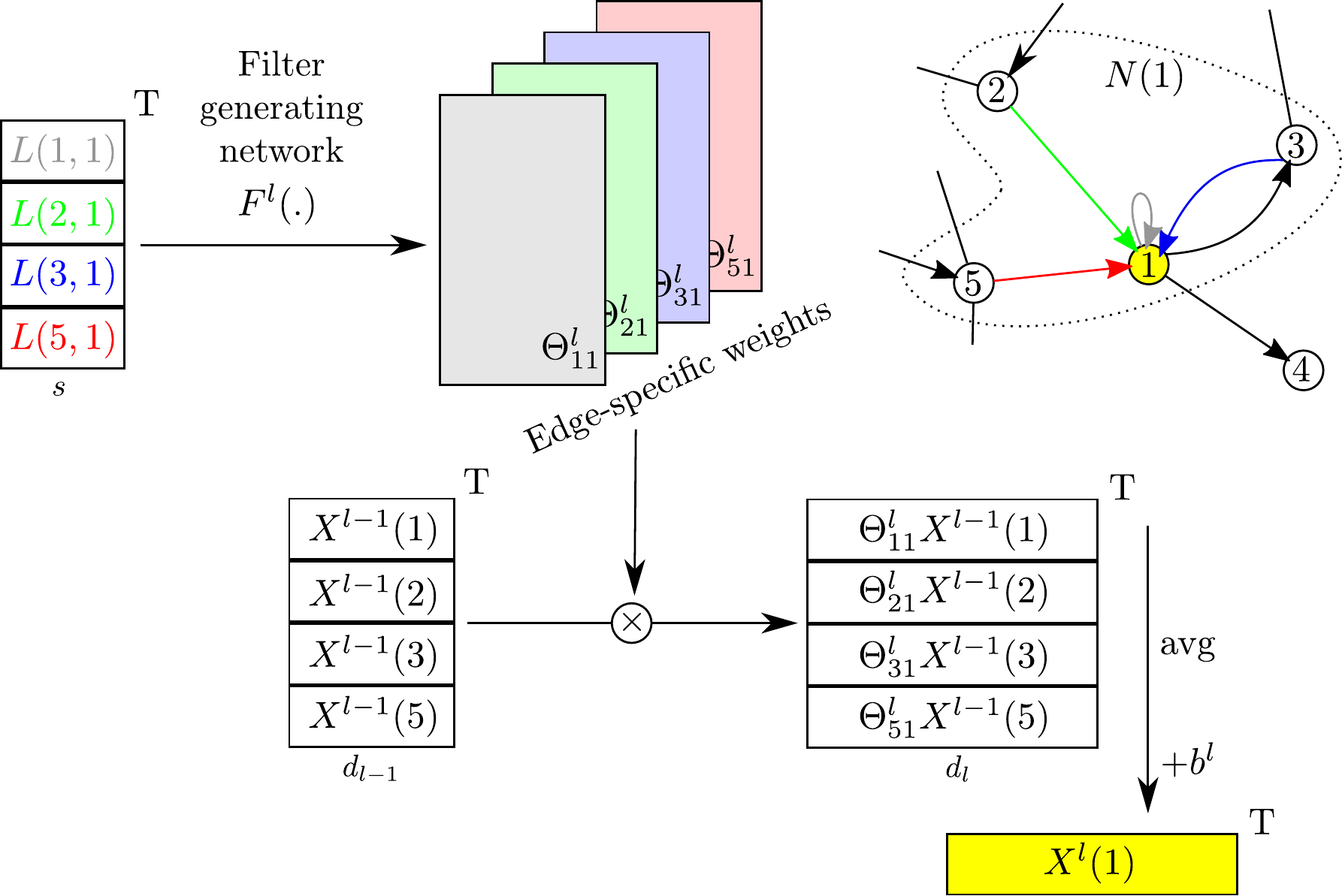}
\vspace{1.5ex}
\caption{\label{fig:gconv} Illustration of edge-conditioned convolution on a directed subgraph. The feature $X^l(1)$ on vertex 1 in the $l$-th network layer is computed as a weighted sum of features $X^{l-1}(.)$ on the set of its predecessor vertices, assuming self-loops. The particular weight matrices are dynamically generated by filter-generating network $F^l$ based on the corresponding edge labels $L(.)$, visualized as colors.}
\end{figure*}

\section{Related Work}

The first formulation of a convolutional network analogy for irregular domains modeled with graphs has been introduced by Bruna \etal~\cite{bruna13}, who looked into both the spatial and the spectral domain of representation for performing localized filtering. 

\paragraph*{Spectral Methods.} A mathematically sound definition of convolution operator makes use of the spectral analysis theory, where it corresponds to multiplication of the signal on vertices transformed into the spectral domain by graph Fourier transform. The spatial locality of filters is then given by smoothness of the spectral filters, in case of \cite{bruna13} modeled as B-splines. The transform involves very expensive multiplications with the eigenvector matrix. However, by a parameterization of filters as Chebyshev polynomials of eigenvalues and their approximate evaluation, computationally efficient and localized filtering has been recently achieved by Defferrard \etal~\cite{defferrard16}. Nevertheless, the filters are still learned in the context of the spectrum of graph Laplacian, which therefore has to be the same for all graphs in a dataset. This means that the graph structure is fixed and only the signal defined on the vertices may differ. This precludes applications on problems where the graph structure varies in the dataset, such as meshes, point clouds, or diverse biochemical datasets.

To cover these important cases, we formulate our filtering approach in the spatial domain, where the limited complexity of evaluation and the localization property is provided by construction. The main challenge here is dealing with weight sharing among local neighborhoods \cite{bruna13}, as the number of vertices adjacent to a particular vertex varies and their ordering is often not well definable.

\paragraph*{Spatial Methods.} Bruna \etal \cite{bruna13} assumed fixed graph structure and did not share any weights among neighborhoods. Several works have independently dealt with this problem. Duvenaud \etal \cite{duvenaud} sum the signal over neighboring vertices followed by a weight matrix multiplication, effectively sharing the same weights among all edges. Atwood and Towsley \cite{dcnn} share weights based on the number of hops between two vertices. Kipf and Welling \cite{kipf} further approximate the spectral method of \cite{defferrard16} and weaken the dependency on the Laplacian, but ultimately arrive at center-surround weighting of neighborhoods. None of these methods captures finer structure of the neighborhood and thus does not generalize the standard convolution on grids. In contrast, our method can make use of possible edge labels and is shown to generalize regular convolution (Section~\ref{subsec:relgridconv}). 

The approach of Niepert \etal \cite{niepert} introduces a heuristic for linearizing selected graph neighborhoods so that a conventional 1D CNN can be used. We share their goal of capturing structure in neighborhoods but approach it in a different way. Finally, Graph neural networks \cite{scarselli09,yujia16} propagate features across a graph until (near) convergence and exploit edge labels as one of the sources of information as we do. However, their system is quite different from the current multilayer feed-forward architectures, making the reuse of today's common building blocks not straightforward.

\paragraph*{CNNs on Point Clouds and Meshes.} There has been little work on deep learning on point clouds or meshes. Masci \etal \cite{masci15} define convolution over patch descriptors around every vertex of a 3D mesh using geodesic distances, formulated in a deep learning architecture. The only way of processing point clouds using deep learning has been to first voxelize them before feeding them to a 3D CNN, be it for classification \cite{voxnet} or segmentation \cite{pclabeling16} purposes. Instead, we regard point cloud as graphs in Euclidean space in this work.

\section{Method} \label{sec:method}

We propose a method for performing convolutions over local graph neighborhoods exploiting edge labels (Section~\ref{subsec:ecc}) and show it to generalize regular convolutions (Section~\ref{subsec:relgridconv}). Afterwards, we present deep networks with our convolution operator (Section~\ref{subsec:eccnet}) in the case of point clouds (Section~\ref{subsec:applclouds}) and general graphs (Section~\ref{subsec:applgraphs}).

\subsection{Edge-Conditioned Convolution} \label{subsec:ecc}

Let us consider a directed or undirected graph $G=(V,E)$, where $V$ is a finite set of vertices with $|V|=n$ and $E\subseteq V\times V$ is a set of edges with $|E|=m$. Let $l\in\{0,..,l_\mathrm{max}\}$ be the layer index in a feed-forward neural network. We assume the graph is both vertex- and edge-labeled, \ie there exists function $X^l:V \mapsto\mathbb{R}^{d_l}$ assigning labels (also called signals or features) to each vertex and $L: E\mapsto\mathbb{R}^s$ assigning labels (also called attributes) to each edge. These functions can be regarded as matrices $X^l\in\mathbb{R}^{n\times d_l}$ and $L\in\mathbb{R}^{m\times s}$, $X^0$ then being the input signal. A neighborhood $N(i)=\{j;(j,i)\in E\}\cup\{i\}$ of vertex $i$ is defined to contain all adjacent vertices (predecessors in directed graphs) including $i$ itself (self-loop).

\def\dimo{{\mathbb{R}^{d_l}}}
\def\dimi{{\mathbb{R}^{d_{l-1}}}}
\def\dimjoint{{\mathbb{R}^{d_l\times d_{l-1}}}}

Our approach computes the filtered signal $X^l(i)\in\dimo$ at vertex $i$ as a weighted sum of signals $X^{l-1}(j)\in\dimi$ in its neighborhood, $j\in N(i)$. While such a commutative aggregation solves the problem of undefined vertex ordering and varying neighborhood sizes, it also smooths out any structural information. To retain it, we propose to condition each filtering weight on the respective edge label. To this end, we borrow the idea from Dynamic filter networks ~\cite{dfn16} and define a filter-generating network $F^l: \mathbb{R}^{s} \mapsto \dimjoint$  which given edge label $L(j,i)$ outputs edge-specific weight matrix $\Theta_{ji}^l \in\dimjoint$, see Figure~\ref{fig:gconv}. 

The convolution operation, coined Edge-Conditioned Convolution (ECC), is formalized as follows:

\begin{equation}
\begin{split}
X^l(i) &= \frac{1}{|N(i)|} \sum_{j\in N(i)} F^l(L(j,i);w^l) X^{l-1}(j) + b^l \label{eq:C1} \\
&= \frac{1}{|N(i)|} \sum_{j\in N(i)} \Theta_{ji}^l X^{l-1}(j) + b^l
\end{split}
\end{equation}

where $b^l\in\dimo$ is a learnable bias and $F^l$ is parameterized by learnable network weights $w^l$. For clarity, $w^l$ and $b^l$ are model parameters updated only during training and $\Theta_{ji}^l$ are dynamically generated parameters for an edge label in a particular input graph. The filter-generating network $F^l$ can be implemented with any differentiable architecture; we use multi-layer perceptrons in our applications.

\paragraph*{Complexity.} Computing $X^l$ for all vertices requires at most\footnote{If edge labels are represented by $s$ discrete values in a particular graph and $s<m$, $F^l$ can be evaluated only $s$-times.} $m$ evaluations of $F^l$ and $m+n$ or $2m+n$ matrix-vector multiplications for directed, resp. undirected graphs. Both operations can be carried out efficiently on the GPU in batch-mode.

\subsection{Relationship to Existing Formulations} \label{subsec:relgridconv}

Our formulation of convolution on graph neighborhoods retains the key properties of the standard convolution on regular grids that are useful in the context of CNNs: weight sharing and locality. 

The weights in ECC are tied by edge label, which is in contrast to tying them by hop distance from a vertex~\cite{dcnn}, according to a neighborhood linearization heuristic~\cite{niepert}, by being the central vertex or not~\cite{kipf}, indiscriminately~\cite{duvenaud}, or not at all~\cite{bruna13}.

In fact, our definition reduces to that of Duvenaud \etal~\cite{duvenaud} (up to scaling) in the case of uninformative edge labels: 
$\sum_{j\in N(i)} \Theta_{ji}^l X^{l-1}(j) = \Theta^l \sum_{j\in N(i)} X^{l-1}(j)$ if $\Theta_{ji}^l = \Theta^l \;\;\forall (j,i)\in E$.

\begin{figure}[bt]
\centering
\includegraphics[width=0.8\linewidth]{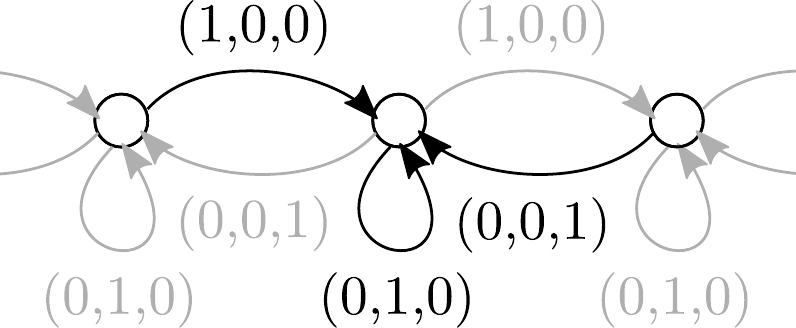}
\vspace{1.5ex}
\caption{\label{fig:regularequiv} Construction of a directed graph with one-hot edge labeling where the proposed edge-conditioned convolution is equivalent to the regular 1D convolution with a centered filter of size $s=3$.}
\end{figure}

More importantly, the standard discrete convolution on grids is a special case of ECC, which we demonstrate in 1D for clarity. Consider an ordered set of vertices $V$ forming a path graph (chain). To obtain convolution with a centered kernel of size $s$, we form $E$ so that each vertex is connected to its $s$ spatially nearest neighbors including self by a directed edge labeled with one-hot encoding of the neighbor's discrete offset $\delta$, see Figure~\ref{fig:regularequiv}. Taking $F^l$ as a single-layer perceptron without bias, we have $F^l(L(j,i);w^l) = w^l(\delta)$, where $w^l(\delta)$ denotes the respective reshaped column of the parameter matrix $w^l\in \mathbb{R}^{(d_l\times d_{l-1})\times s}$. With a slight abuse of notation, we arrive at the equivalence to the standard convolution: $X^l(i) = \sum_{j\in N(i)} \Theta_{ji}^l X^{l-1}(j) = \sum_{\delta} w^l(\delta) X^{l-1}(i-\delta)$, ignoring the normalization factor of $1/|N(i)|$ playing a role only at grid boundaries.

This shows that ECC can retain the same number of parameteres and computational complexity of the regular convolution in the case of grids. Note that such equivalence is not possible with none of \cite{dcnn,kipf,duvenaud} due to their way of weight tying.

\subsection{Deep Networks with ECC}  \label{subsec:eccnet}

\begin{figure*}[bt]
\centering
\includegraphics[width=0.75\linewidth]{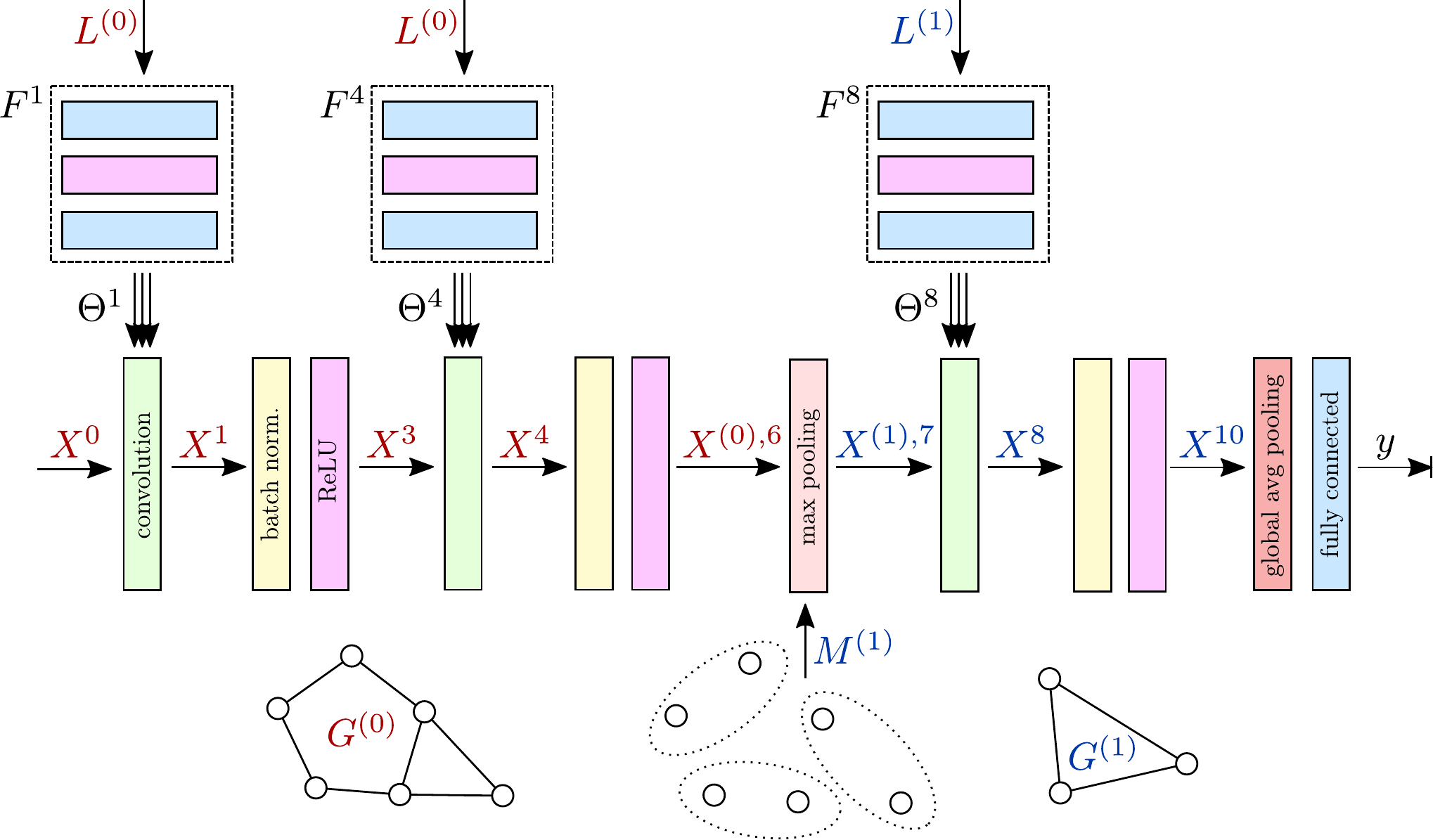}
\vspace{1.5ex}
\caption{\label{fig:netexample} Illustration of a deep network with three edge-conditioned convolutions (first, fourth, and eight network layer) and one pooling (seventh layer). The last convolution is executed on a structurally different graph $G^{(1)}$, which is related to the input graph $G^{(0)}$ by coarsening and signal aggregation in the max pooling step according to mapping $M^{(1)}$. See Section~\ref{subsec:eccnet} for more details.}
\end{figure*}

While ECC is in principle applicable to both vertex classification and graph classification tasks, in this paper we restrict ourselves only to the latter one, \ie predicting a class for the whole input graph. Hence, we follow the common architectural pattern for feed-forward networks of interlaced convolutions and poolings topped by global pooling and fully-connected layers, see Figure~\ref{fig:netexample} for an illustration. This way, information from the local neighborhoods gets combined over successive layers to gain context (enlarge receptive field). While edge labels are fixed for a particular graph, their (learned) interpretation by the means of filter generating networks may change from layer to layer (weights of $F^l$ are not shared among layers). Therefore, the restriction of ECC to 1-hop neighborhoods $N(i)$ is not a constraint, akin to using small 3$\times$3 filters in normal CNNs in exchange for deeper networks, which is known to be beneficial~\cite{hesun14}.

We use batch normalization~\cite{batchnorm} after each convolution, which was necessary for the learning to converge. Interestingly, we had no success with other feature normalization techniques such as data-dependent initialization~\cite{goodinit} or layer normalization~\cite{layernorm}.

\paragraph*{Pooling.} While (non-strided) convolutional layers and all point-wise layers do not change the underlying graph and only evolve the signal on vertices, pooling layers are defined to output aggregated signal on the vertices of a new, coarsened graph. Therefore, a pyramid of $h_\mathrm{max}$ progressively coarser graphs has to be constructed for each input graph. Let us extend here our notation with an additional superscript $h\in\{0,..,h_\mathrm{max}\}$ to distinguish among different graphs $G^{(h)}=(V^{(h)}, E^{(h)})$ in the pyramid when necessary. Each $G^{(h)}$ has also its associated labels $L^{(h)}$ and signal $X^{{(h)},l}$.
A coarsening typically consists of three steps: subsampling or merging vertices, creating the new edge structure $E^{(h)}$ and labeling $L^{(h)}$ (so-called reduction), and mapping the vertices in the original graph to those in the coarsened one with $M^{(h)}: V^{(h-1)}\mapsto V^{(h)}$. We use a different algorithm depending on whether we work with general graphs or graphs in Euclidean space, therefore we postpone discussing the details to the applications. Finally, the pooling layer with index $l_h$ aggregates $X^{(h-1),l_h-1}$ into a lower dimensional $X^{(h),l_h}$ based on $M^{(h)}$. See Figure~\ref{fig:netexample} for an example of using the introduced notation. 

During coarsening, a small graph may be reduced to several disconnected vertices in its lower resolutions without problems as self-edges are always present. Since the architecture is designed to process graphs with variable $n,m$, we deal with varying vertex count $n^{(h_{max})}$ in the lowest graph resolution by global average/max pooling.

\subsection{Application in Point Clouds} \label{subsec:applclouds}

Point clouds are an important 3D data modality arising from many acquisition techniques, such as laser scanning (LiDAR) or multi-view reconstruction. Due to their natural irregularity and sparsity, so far the only way of processing point clouds using deep learning has been to first voxelize them before feeding them to a 3D CNN, be it for classification~\cite{voxnet} or segmentation~\cite{pclabeling16} purposes. Such a dense representation is very hardware friendly and simple to handle with the current deep learning frameworks.

On the other hand, there are several disadvantages too. First, voxel representation tends to be much more expensive in terms of memory than usually sparse point clouds (we are not aware of any GPU implementation of convolutions on sparse tensors). Second, the necessity to fit them into a fixed size 3D grid brings about discretization artifacts and the loss of metric scale and possibly of details. With this work, we would like to offer a competitive alternative to the mainstream by performing deep learning on point clouds directly. As far as we know, we are the first to demonstrate such a result.

\paragraph*{Graph Construction.} Given a point cloud $P$ with its point features $X_P$ (such as laser return intensity or color) we build a directed graph $G=(V,E)$ and set up its labels $X^0$ and $L$ as follows. First, we create vertex $i\in V$ for every point $p\in P$ and assign the respective signal to it by $X^0(i)=X_P(p)$ (or 0 if there are no features $X_P(p)$). Then we connect each vertex $i$ to all vertices $j$ in its spatial neighborhood by a directed edge $(j,i)$. In our experiments with neighborhoods, fixed metric radius $\rho$ worked better than a fixed number of neighbors. 
The offset $\delta=p_j-p_i$ between the points corresponding to vertices $j$, $i$ is represented in Cartesian and spherical coordinates as 6D edge label vector $L(j,i) = (\delta_x, \delta_y, \delta_z, ||\delta||, \arccos \delta_z/||\delta||, \arctan \delta_y/\delta_x)$.

\paragraph*{Graph Coarsening.} For a single input point cloud $P$, a pyramid of downsampled point clouds $P^{(h)}$ is obtained by the VoxelGrid algorithm~\cite{pclrusu}, which overlays a grid of resolution $r^{(h)}$ over the point cloud and replaces all points within a voxel with their centroid (and thus maintains subvoxel accuracy). Each of the resulting point clouds $P^{(h)}$ is then independently converted into a graph $G^{(h)}$ and labeling $L^{(h)}$ with neighborhood radius $\rho^{(h)}$ as described above. The pooling map $M^{(h)}$ is defined so that each point in $P^{(h-1)}$ is assigned to its spatially nearest point in the subsampled point cloud $P^{(h)}$.

\paragraph*{Data Augmentation.} In order to reduce overfitting on small datasets, we perform online data augmentation. In particular, we randomly rotate point clouds about their up-axis, jitter their scale, perform mirroring, or delete random points.

\subsection{Application in General Graphs} \label{subsec:applgraphs}

Many problems can be modeled directly as graphs. In such cases the graph dataset is already given and only the appropriate graph coarsening scheme needs to be chosen. This is by no means trivial and there exists a large body of literature on this problem~\cite{safro14}. Without any concept of spatial localization of vertices, we resort to established graph coarsening algorithms and utilize the multiresolution framework of Shuman \etal~\cite{shumanFV16,gspbox}, which works by repeated downsampling and graph reduction of the input graph. The downsampling step is based on splitting the graph into two components by the sign of the largest eigenvector of the Laplacian. This is followed by Kron reduction~\cite{kron}, which also defines the new edge labeling, enhanced with spectral sparsification of edges~\cite{spielman2011graph}. Note that the algorithm regards graphs as unweighted for the purpose of coarsening.

This method is attractive for us because of two reasons. Each downsampling step removes approximately half of the vertices, guaranteeing a certain level of pooling strength, and the sparsification step is randomized. The latter property is exploited as a useful data augmentation technique since several different graph pyramids can be generated from a single input graph. This is in spirit similar to the effect of fractional max-pooling~\cite{fractpool}. We do not perform any other data augmentation.

\section{Experiments}

The proposed method is evaluated in point cloud classification (real-world data in Section~\ref{subsec:evalpc} and synthetic in \ref{subsec:evalmn}) and on a standard graph classification benchmark (Section~\ref{subsec:evalgg}). In addition, we validate our method and study its properties on MNIST (Section~\ref{subsec:evalmnist}).

\subsection{Sydney Urban Objects} \label{subsec:evalpc}

\begin{figure}[bt]
\centering
\includegraphics[width=\linewidth]{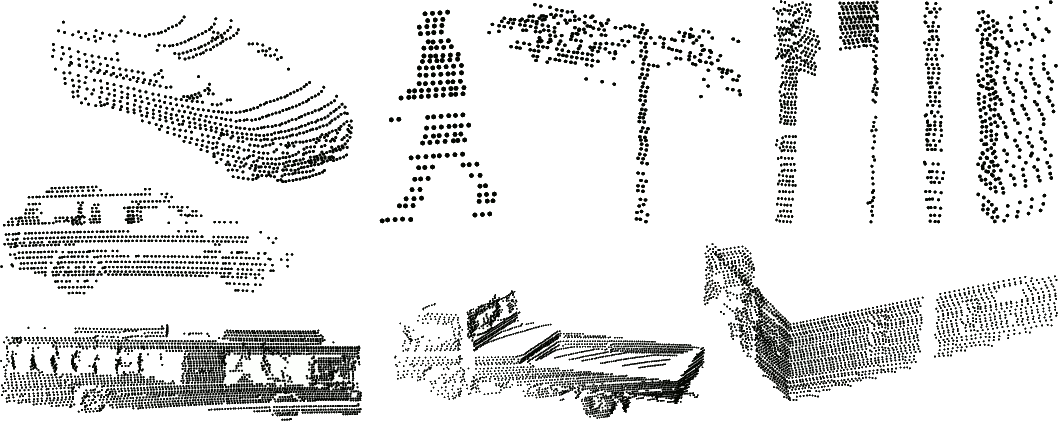}
\vspace{1.5ex}
\caption{\label{fig:sydney} Illustrative samples of the majority of classes in Sydney Urban Objects dataset, reproduced from \cite{trianglesvm}.}
\end{figure}

This point cloud dataset \cite{trianglesvm} consists of 588 objects in 14 categories (vehicles, pedestrians, signs, and trees) manually extracted from 360$^{\circ}$ LiDAR scans, see Figure~\ref{fig:sydney}. It demonstrates non-ideal sensing conditions with occlusions (holes) and a large variability in viewpoint (single viewpoint). This makes object classification a challenging task. 

Following the protocol employed by the dataset authors, we report the mean F1 score weighted by class frequency, as the dataset is imbalanced. This score is further aggregated over four standard training/testing splits.

\paragraph*{Network Configuration.}
Our ECC-network has 7 parametric layers and 4 levels of graph resolution. Its configuration can be described as C(16)-C(32)-MP(0.25,0.5)-C(32)-C(32)-MP(0.75,1.5)-C(64)-MP(1.5,1.5)-GAP-FC(64)-D(0.2)-FC(14), where C($c$) denotes ECC with $c$ output channels followed by affine batch normalization and ReLU activation, MP($r$,$\rho$) stands for max-pooling down to grid resolution of $r$ meters and neighborhood radius of $\rho$ meters, GAP is global average pooling, FC($c$) is fully-connected layer with $c$ output channels, and D($p$) is dropout with probability $p$. The filter-generating networks $F^l$ have configuration FC(16)-FC(32)-FC($d_l d_{l-1}$) with orthogonal weight initialization~\cite{orthoinit} and ReLUs in between. Input graphs are created with $r^0=0.1$ and $\rho^0=0.2$ meters to break overly dense point clusters. Networks are trained with SGD and cross-entropy loss for 250 epochs with batch size 32 and learning rate 0.1 step-wise decreasing after 200 and 245 epochs. Vertex signal $X^0$ is scalar laser return intensity (0-255), representing depth.

\paragraph*{Results.} 
Table~\ref{tab:respc} compares our result (ECC, 78.4) against two methods based on volumetric CNNs evaluated on voxelized occupancy grids of size 32x32x32 (VoxNet \cite{voxnet} 73.0 and ORION \cite{orion} 77.8), which we outperform by a small margin and set the new state of the art result on this dataset. 

In the same table, we also study the dependence on convolution radii $\rho$: increasing them $1.5\times$ or $2\times$ in all convolutional layers leads to a drop in performance, which would correspond to a preference of using smaller filters in regular CNNs. The average neighborhood size is roughly 10 vertices for our best-performing network.
We hypothesize that larger radii smooth out the information in the central vertex. 
To investigate this, we increased the importance of the self-loop by adding an identity skip-connection (see Appendix~\ref{sec:idconn}) and retrained the networks. We achieved 77.0, 79.5 (the new state of the art), and 77.4 mean F1 for ECC, ECC $1.5\rho$, and ECC $2\rho$, respectively. Stronger identity connection allowed for successful integration of a larger context, up to some limit, which indeed suggests that information should be aggregated neither too much nor too little.

\begin{table}[bt]
\centering
\begin{tabular}{cc}
\toprule
Model & Mean F1\tabularnewline
\midrule
Triangle+SVM \cite{trianglesvm}& 67.1 \tabularnewline
GFH+SVM \cite{gfhsvm} & 71.0 \tabularnewline
VoxNet \cite{voxnet} & 73.0 \tabularnewline
ORION \cite{orion} & 77.8\tabularnewline
\midrule
ECC $2\rho$ & 74.4 \tabularnewline
ECC $1.5\rho$ & 76.9 \tabularnewline
ECC & 78.4\tabularnewline
\bottomrule
\end{tabular}
\vspace{1.5ex}
\caption{\label{tab:respc}
Mean F1 score weighted by class frequency on Sydney Urban Objects dataset \cite{trianglesvm}. Only the best-performing models of each baseline are listed.}
\end{table}

\subsection{ModelNet} \label{subsec:evalmn}

ModelNet~\cite{modelnet} is a large scale collection of object meshes. We evaluate classification performance on its subsets ModelNet10 (3991/908 train/test examples in 10 categories) and ModelNet40 (9843/2468 train/test examples in 40 categories). Synthetic point clouds are created from meshes by uniformly sampling 1000 points on mesh faces according to face area (a simulation of acquisition from multiple viewpoints) and rescaled into a unit sphere.

\paragraph*{Network Configuration.}
Our ECC-network for ModelNet10 has 7 parametric layers and 3 levels of graph resolution with configuration C(16)-C(32)-MP(2.5/32,7.5/32)-C(32)-C(32)-MP(7.5/32,22.5/32)-C(64)-GMP-FC(64)-D(0.2)-FC(10), GMP being global max pooling. Other definitions and filter-generating networks $F^l$ are as in Section~\ref{subsec:evalpc}. Input graphs are created with $r^0=1/32$ and $\rho^0=2/32$ units, mimicking the typical grid resolution of $32^3$ in voxel-based methods. The network is trained with SGD and cross-entropy loss for 175 epochs with batch size 64 and learning rate 0.1 step-wise decreasing after every 50 epochs. There is no vertex signal, \ie $X^0$ are zero. For ModelNet40, the network is wider (C(24), C(48), C(48), C(48), C(96), FC(64), FC(40)) and is trained for 100 epochs with learning rate decreasing after each 30 epochs.

\paragraph*{Results.} 
Table~\ref{tab:resmn} compares our result to several recent works, based either on volumetric \cite{modelnet, voxnet, orion, qi16} or rendered image representation \cite{Su15}. Test sets were expanded to include 12 orientations (ECC). We also evaluate voting over orientations (ECC 12 votes), which slightly improves the results likely due to the rotational variance of VoxelGrid algorithm. While not fully reaching the state of the art, we believe our method remains very competitive (90.8\%, resp. 87.4\% mean instance accuracy). For a fairer comparison, a leading volumetric method should be retrained on voxelized synthetic point clouds.

\begin{table}[bt]
\centering
\begin{tabular}{ccc}
\toprule
Model & ModelNet10 & ModelNet40 \tabularnewline
\midrule
3DShapeNets \cite{modelnet} & 83.5 & 77.3 \tabularnewline
MVCNN \cite{Su15} & {\textemdash} & 90.1 \tabularnewline
VoxNet \cite{voxnet} & 92 & 83 \tabularnewline
ORION \cite{orion} & 93.8 & {\textemdash} \tabularnewline
SubvolumeSup \cite{qi16} & {\textemdash} & 86.0 (89.2) \tabularnewline
\midrule
ECC & 89.3 (90.0) & 82.4 (87.0) \tabularnewline
ECC (12 votes) & 90.0 (90.8) & 83.2 (87.4)\tabularnewline
\bottomrule
\end{tabular}
\vspace{1.5ex}
\caption{\label{tab:resmn}
Mean class accuracy (resp. mean instance accuracy) on ModelNets~\cite{modelnet}. Only the best models of each baseline are listed.}
\end{table}

\subsection{Graph Classification} \label{subsec:evalgg}

We evaluate on a graph classification benchmark frequently used in the community, consisting of five datasets: NCI1, NCI109, MUTAG, ENZYMES, and D{\&}D. Their properties can be found in Table~\ref{tab:stats}, indicating the variability in dataset sizes, in graph sizes, and in the availability of labels. Following \cite{shervashidze}, we perform 10-fold cross-validation with 9 folds for training and 1 for testing and report the average prediction accuracy.

NCI1 and NCI109~\cite{nci1db} consist of graph representations of chemical compounds screened for activity against non-small cell lung cancer and ovarian cancer cell lines, respectively. MUTAG~\cite{debnath1991structure} is a dataset of nitro compounds labeled according to whether or not they have a mutagenic effect on a bacterium. ENZYMES~\cite{borgwardtK05} contains representations of tertiary structure of 6 classes of enzymes. D{\&}D~\cite{dobson2003} is a database of protein structures (vertices are amino acids, edges indicate spatial closeness) classified as enzymes and non-enzymes. 

\paragraph*{Network Configuration.}
Our ECC-network for NCI1 has 8 parametric layers and 3 levels of graph resolution. Its configuration can be described as C(48)-C(48)-C(48)-MP-C(48)-C(64)-MP-C(64)-GAP-FC(64)-D(0.1)-FC(2), where C($c$) denotes ECC with $c$ output channels followed by affine batch normalization, ReLU activation and dropout (probability 0.05), MP stands for max-pooling onto a coarser graph, GAP is global average pooling, FC($c$) is fully-connected layer with $c$ output channels, and D($p$) is dropout with probability $p$. The filter-generating networks $F^l$ have configuration FC(64)-FC($d_l d_{l-1}$) with orthogonal weight initialization~\cite{orthoinit} and ReLU in between. Labels are encoded as one-hot vectors ($d_0=37$ and $s=4$ due to an extra label for self-connections). Networks are trained with SGD and cross-entropy loss for 50 epochs with batch size 64 and learning rate 0.1 step-wise decreasing after 25, 35, and 45 epochs. The dataset is expanded five times by randomized sparsification (Section~\ref{subsec:applgraphs}). Small deviations from this description for the other four datasets are mentioned in the supplementary.

\paragraph*{Baselines.}
We compare our method (ECC) to the state of the art Weisfeiler-Lehman graph kernel \etal \cite{shervashidze} and to four approaches using deep learning as at least one of their components \cite{dcnn, niepert, deepkern, struct2vec}. Randomized sparsification used during training time can also be exploited at test time, when the network prediction scores (ECC-5-scores) or votes (ECC-5-votes) are averaged over 5 runs. To judge the influence of edge labels, we run our method with uniform labels and $F^l$ being a single layer FC($d_l d_{l-1}$) without bias\footnote{Also possible for unlabeled ENZYMES and D{\&}D, since our method uses labels from Kron reduction for all coarsened graphs by default.} (ECC no edge labels). 

\paragraph*{Results.}
Table~\ref{tab:resgraph} conveys that while there is no clear winning algorithm, our method performs at the level of state of the art for edge-labeled datasets (NCI1, NCI109, MUTAG). The results demonstrate the importance of exploiting edge labels for convolution-based methods, as the performance of DCNN \cite{dcnn} and ECC without edge labels is distinctly worse, justifying the motivation behind this paper. Averaging over random sparsifications at test time improves accuracy by a small amount. Our results on datasets without edge labels (ENZYMES, D{\&}D) are somewhat below the state of the art but still at a reasonable level, though improvement in this case was not the aim of this work. This indicates that further research is needed into the adaptation of CNNs to general graphs. A more detailed discussion for each dataset is available in the supplementary.

\begin{table}[bt]
\centering
\addtolength{\tabcolsep}{-3pt}
\resizebox{1\linewidth}{!}{
\begin{tabular}{cccccc}
\toprule
 & NCI1 & NCI109 & MUTAG & ENZYMES & D{\&}D \tabularnewline
\midrule
\# graphs & 4110 & 4127 & 188 & 600 & 1178\tabularnewline
mean $|V|$ & 29.87 & 29.68 & 17.93 & 32.63 & 284.32 \tabularnewline
mean $|E|$ & 32.3 & 32.13 & 19.79 & 62.14 & 715.66 \tabularnewline
\# classes & 2 & 2 & 2 & 6 & 2\tabularnewline
\# vertex labels & 37 & 38 & 7 & 3 & 82\tabularnewline
\# edge labels & 3 & 3 & 11 & {\textemdash} & {\textemdash}\tabularnewline
\bottomrule
\end{tabular}}
\addtolength{\tabcolsep}{3pt}
\vspace{1.5ex}
\caption{\label{tab:stats}
Characteristics of the graph benchmark datasets, extended from \cite{struct2vec}. Both edge and vertex labels are categorical.}
\end{table}

\begin{table}[bt]
\centering
\addtolength{\tabcolsep}{-3pt}
\resizebox{1\linewidth}{!}{
\begin{tabular}{cSSSSS}
\toprule
Model & {NCI1} & {NCI109} & {MUTAG} & {ENZYMES} & {D{\&}D} \tabularnewline
\midrule
DCNN \cite{dcnn} & 62.61 & 62.86 & 66.98 & 18.10 & {\textemdash} \tabularnewline
PSCN \cite{niepert} & 78.59 & {\textemdash} & 92.63 & {\textemdash} & 77.12 \tabularnewline
Deep WL \cite{deepkern} & 80.31 & 80.32 & 87.44 & 53.43 & {\textemdash} \tabularnewline
structure2vec \cite{struct2vec} & 83.72 & 82.16 & 88.28 & 61.10 & 82.22\tabularnewline
WL \cite{shervashidze} & 84.55 & 84.49 & 83.78 & 59.05 & 79.78\tabularnewline
\midrule
ECC (no edge labels)& 76.82 & 75.03 & 76.11 & 45.67 & 72.54 \tabularnewline
ECC 				& 83.80 & 81.87 & 89.44 & 50.00 & 73.65 \tabularnewline
ECC (5 votes) 		& 83.63 & 82.04 & 88.33 & 53.50 & 73.68 \tabularnewline
ECC (5 scores)		& 83.80 & 82.14 & 88.33 & 52.67 & 74.10 \tabularnewline
\bottomrule
\end{tabular}}
\addtolength{\tabcolsep}{3pt}
\vspace{1.5ex}
\caption{\label{tab:resgraph}
Mean accuracy (10 folds) on graph classification datasets. Only the best-performing models of each baseline are listed.}
\end{table}

\subsection{MNIST} \label{subsec:evalmnist}

To further validate our method, we applied it to the MNIST classification problem \cite{mnist}, a dataset of 70k greyscale images of handwritten digits represented on a 2D grid of size 28$\times$28. We regard each image $I$ as point cloud $P$ with points $p_i=(x,y,0)$ and signal $X^0(i)=I(x,y)$ representing each pixel, $x,y\in\{0,..,27\}$. Edge labeling and graph coarsening is performed as explained in Section~\ref{subsec:applclouds}. 
We are mainly interested in two questions: Is ECC able to reach the standard performance on this classic baseline? What kind of representation does it learn?

\paragraph*{Network Configuration.}
Our ECC-network has 5 parametric layers with configuration C(16)-MP(2,3.4)-C(32)-MP(4,6.8)-C(64)-MP(8,30)-C(128)-D(0.5)-FC(10); the notation and filter-generating network being as in Section~\ref{subsec:evalpc}. The last convolution has a stride of 30 and thus maps all $4\times 4$ points to only a single point. Input graphs are created with $r^0=1$ and $\rho^0=2.9$. This model exactly corresponds to a regular CNN with three convolutions with filters of size 5$\times$5, 3$\times$3, and 3$\times$3 interlaced with max-poolings of size 2$\times$2, finished with two fully connected layers. Networks are trained with SGD and cross-entropy loss for 20 epochs with batch size 64 and learning rate 0.01 step-wise decreasing after 10 and 15 epochs.

\paragraph*{Results.} Table~\ref{tab:resmnist} proves that our ECC network can achieve the level of quality comparable to the good standard in the community (99.14). This is exactly the same accuracy as reported by Defferrard \etal~\cite{defferrard16} and better than what is offered by other spectral-based approaches (98.2 \cite{bruna13}, 94.96 \cite{edwards16}). Note that we are not aiming at becoming the state of the art on MNIST by this work. 

Next, we investigate the effect of regular grid and irregular mesh. To this end, we discard all black points ($X^0(i)=0$) from the point clouds, corresponding to 80.9\% of data, and retrain the network (ECC sparse input). Exactly the same test performance is obtained (99.14), indicating that our method is very stable with respect to graph structure changing from sample to sample.

Furthermore, we check the quality of the learned filter generating networks $F^l$. We compare with ECC configured to mimic regular convolution using single-layer filter networks and one-hot encoding of offsets (ECC one-hot), as described in Section~\ref{subsec:relgridconv}. This configuration reaches 99.37 accuracy, or 0.23 more than ECC, implying that $F^l$ are not perfect but still perform very well in learning the proper partitioning of edge labels.

\begin{figure}[bt]
\centering
\includegraphics[width=0.49\linewidth]{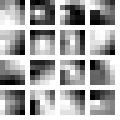}
\includegraphics[width=0.49\linewidth]{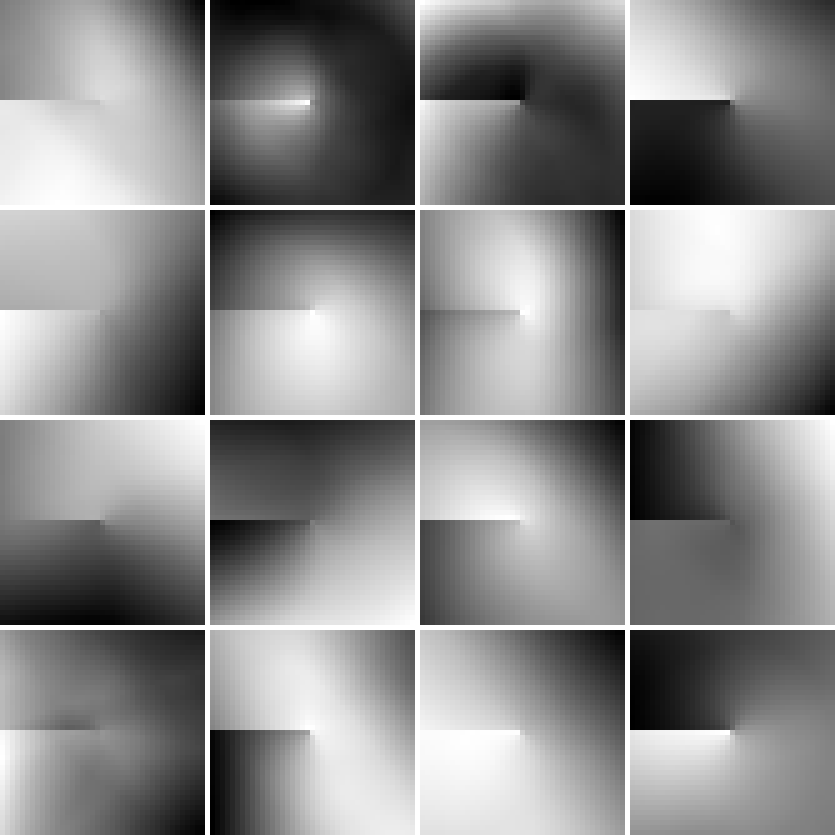}
\vspace{1.5ex}
\caption{\label{fig:filters} Convolutional filters learned on MNIST in the first layer for sparse input ECC, sampled in two different resolutions. See Section~\ref{subsec:evalmnist} for details.}
\end{figure}

Last, we explore the generated filters visually for the case of the sparse input ECC. As filters $\Theta^1 \in \mathbb{R}^{16\times 1}$ are a continuous function of an edge label, we can visualize the change of values in each dimension in 16 images by sampling labels over grids of two resolutions. The coarser one in Figure~\ref{fig:filters} has integer steps corresponding to the offsets $\delta_x,\delta_y\in\{-2,..,2\}$. It shows filters exhibiting the structured patterns typically found in the first layer of CNNs. The finer resolution in Figure~\ref{fig:filters} (sub-pixel steps of 0.1) reveals that the filters are in fact smooth and do not contain any discontinuities apart from the angular artifact due to the $2\pi$ periodicity of azimuth. Interestingly, the artifact is not distinct in all filters, suggesting the network may learn to overcome it if necessary.

\begin{table}[bt]
\centering
\begin{tabular}{ccc}
\toprule
Model & Train accuracy & Test accuracy\tabularnewline
\midrule
ECC & 99.12 & 99.14 \tabularnewline
ECC (sparse input) & 99.36 & 99.14 \tabularnewline
ECC (one-hot) & 99.53 & 99.37\tabularnewline
\bottomrule
\end{tabular}
\vspace{1.5ex}
\caption{\label{tab:resmnist} Accuracy on MNIST dataset \cite{mnist}.}
\end{table}

\section{Conclusion}

We have introduced edge-conditioned convolution (ECC), an operation on graph signal performed in the spatial domain where filter weights are conditioned on edge labels and dynamically generated for each specific input sample. We have shown that our formulation generalizes the standard convolution on graphs if edge labels are chosen properly and experimentally validated this assertion on MNIST. We applied our approach to point cloud classification in a novel way, setting a new state of the art performance on Sydney dataset. Furthermore, we have outperformed other deep learning-based approaches on graph classification dataset NCI1. The source code is available at \url{https://github.com/mys007/ecc}.

In feature work we would like to treat meshes as graphs rather than point clouds. Moreover, we plan to address the currently higher level of GPU memory consumption in case of large graphs with continuous edge labels, for example by randomized clustering, which could also serve as additional regularization through data augmentation.

\subsubsection*{Acknowledgments.} We gratefully acknowledge NVIDIA Corporation for the donated GPU used in this research. We are thankful to anonymous reviewers for their feedback.

{\small
\bibliographystyle{ieee}
\bibliography{pcconv}
}


\appendix
\section*{Appendix} 

\section{Overview}

In the first part, the appendix provides further discussion of the graph classification results (Section \ref{sec:gcdetails}) and investigates robustness of point cloud classification to noise (Section \ref{sec:pcnoise}). In the second part, we explore several extensions of our ECC formulation, specifically with different edge labeling for point clouds (Section \ref{sec:pclabels}), with identity connections (Section \ref{sec:idconn}), with degree labels (Section \ref{sec:degrees}), and with a learned normalization factor  (Section \ref{sec:normnet}).

\section{Details on Graph Classification Benchmark} \label{sec:gcdetails}

In this section we describe the differences in our network architecture to the one introduced for NCI1 in the main paper and discuss evaluation results for each dataset in detail.

\paragraph*{NCI1.} ECC (83.80\%) performs distinctly better than convolution methods that are not able to use edge labels (DCNN \cite{dcnn} 62.61\%, PSCN \cite{niepert} 78.59\%). Methods not approaching the problem as convolutions on graphs but rather combining deep learning with other techniques are stronger (Deep WL \cite{deepkern} 80.31\%, structure2vec \cite{struct2vec} 83.72\%) but are still outperformed by ECC. While the Weisfeiler-Lehman graph kernel remains the strongest method (WL \cite{shervashidze} 84.55\%), it is fair to conclude that ECC, structure2vec, and WL perform at the same level.

\paragraph*{NCI109.} We use the same ECC-network configuration and training details as described in Section~\ref{subsec:evalgg} for NCI1, since both datasets are similar. ECC (82.14\%) performs distinctly better than DCNN \cite{dcnn} (62.86\%), which is not able to use edge labels, and is on par with non-convolutional approaches (Deep WL \cite{deepkern} 80.32\%, structure2vec \cite{struct2vec} 82.16\%, WL \cite{shervashidze} 84.49\%).

\paragraph*{MUTAG.} As MUTAG is a tiny dataset of small graphs, we trained a downsized ECC-network to combat overfitting. Using the notation from Section~\ref{subsec:evalgg}, its configuration is C(16)-C(32)-C(48)-MP-C(64)-MP-GAP-FC(64)-D(0.2)-FC(2), all other details are as with NCI1. While by numbers ECC (89.44\%) outperforms all other approaches except of PSCN \cite{niepert} (92.63\%), we note that all four leading methods (Deep WL \cite{deepkern} 87.44\%, structure2vec \cite{struct2vec} 88.28\%, ECC, PSCN) can be seen to perform equally well due to fluctuations caused by the dataset size. We account the tiny decrease in performance with test-time randomization (88.33\%) to the same reason.

\paragraph*{ENZYMES.} Due to higher complexity of this task we use a wider ECC-network configured as C(64)-C(64)-C(96)-MP-C(96)-C(128)-MP-C(128)-C(160)-MP-C(160)-GAP-FC(192)-D(0.2)-FC(6) using the notation and other details in Section~\ref{subsec:evalgg}. As this dataset is not edge-labeled, we do not expect to obtain the best performance. Indeed, our method (53.50\%) performs at the level of Deep WL \cite{deepkern} (53.43\%) and is overperformed by WL \cite{shervashidze} (59.05\%) and structure2vec \cite{struct2vec} (61.10\%). Note that the gap to the other convolution-based method DCNN \cite{dcnn} (18.10\%) is huge and there is an improvement of more than 4 percentage points due to edge labels in coarser graph resolutions from Kron reduction.

\paragraph*{D{\&}D.} Due to large graphs in this dataset we designed a ECC-network with more pooling configured as C(48)-C(48)-C(48)-MP-C(48)-MP-C(64)-MP-C(64)-MP-C(64)-MP-C(64)-MP-GAP-FC(64)-D(0.2)-FC(2) using the notation and other details in Section~\ref{subsec:evalgg}. As this dataset is not edge-labeled, we do not expect to obtain the best performance. Our method (74.10\%) is overperformed by the others who evaluated on this dataset (PSCN \cite{niepert} 77.12\%, WL \cite{shervashidze} 79.78\%, structure2vec \cite{struct2vec} 82.22\%), though the margin is not very large.

\section{Robustness to Noise} \label{sec:pcnoise}

\begin{figure}[bt]
\centering
\includegraphics[width=0.8\columnwidth]{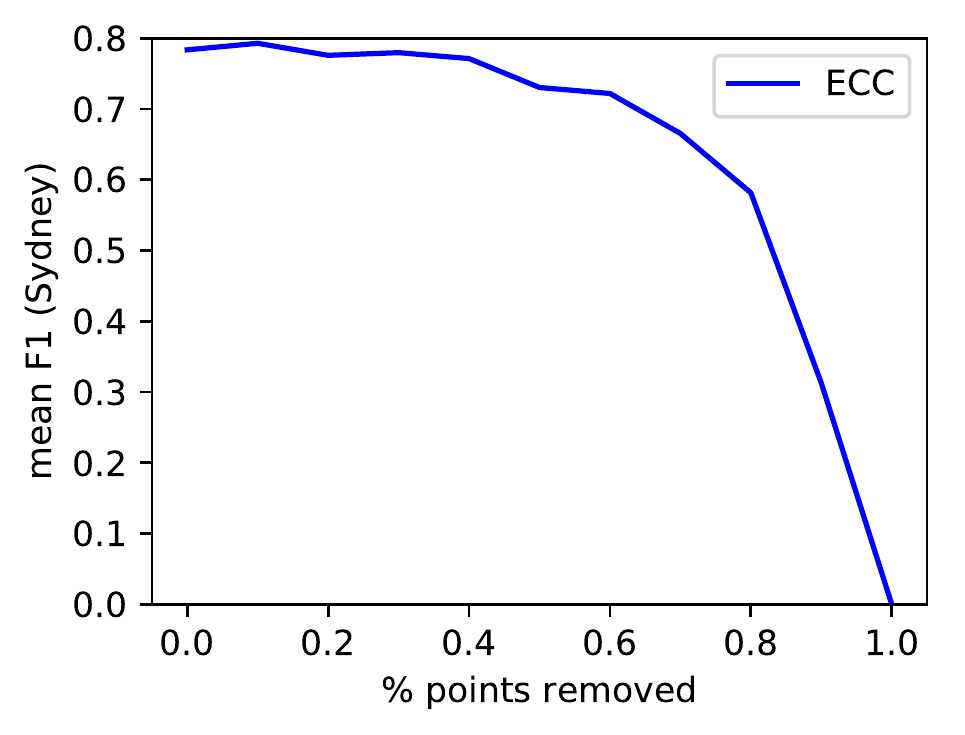} \\
\includegraphics[width=0.8\columnwidth]{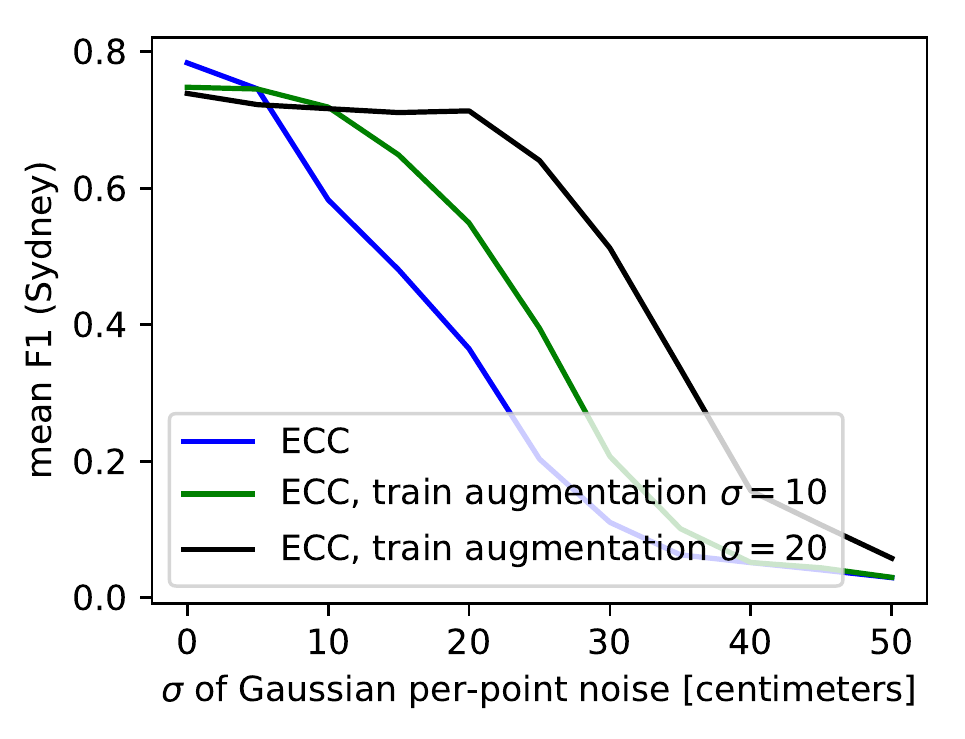}
\caption{\label{fig:robutness}
Robustness to point removal and Gaussian noise.}
\end{figure}

Real-world point clouds contain several kinds of artifacts, such as holes due to occlusions and Gaussian noise due to measurement uncertainty. Figure~\ref{fig:robutness} shows that ECC is highly robust to point removal and can be made robust to additive Gaussian noise by a proper training data augmentation.

\section{Edge Labels for Point Clouds} \label{sec:pclabels}

In Section \ref{subsec:applclouds} we defined edge labels $L(j,i)$ as the offset $\delta=p_j-p_i$ in Cartesian and spherical coordinates, $L(j,i) = (\delta_x, \delta_y, \delta_z, ||\delta||, \arccos \delta_z/||\delta||, \arctan \delta_y/\delta_x)$. Here, we explore the importance of individual elements in the proposed edge labeling and further evaluate labels invariant to rotation about objects' vertical axis $z$ (IRz). Table~\ref{tab:edgelimp} conveys that models with isotropic (60.7) or no labels (38.9) perform poorly as expected, while either of the coordinate systems is important. IRz labeling performs comparably or even slightly better than our proposed one. However, we believe this is a property of the specific dataset and may not necessarily generalize, an example being MNIST, where IRz is equivalent to full isotropy and decreases accuracy to 89.9\%.

\begin{table}[bt]
\centering
\resizebox{1\columnwidth}{!}{%
\begin{tabular}{cc}
\toprule
Label $L(j,i)$ & Mean F1\tabularnewline
\midrule
$(\delta_x, \delta_y, \delta_z, ||\delta||, \arccos \delta_z/||\delta||, \arctan \delta_y/\delta_x)$ & 78.4 \tabularnewline
$(\delta_x, \delta_y, \delta_z)$ & 76.1 \tabularnewline
$(||\delta||, \arccos \delta_z/||\delta||, \arctan \delta_y/\delta_x)$ & 77.3 \tabularnewline
\midrule 
$(||\delta_{xy}||, \delta_z, ||\delta||, \arccos \delta_z/||\delta||)$ & 75.8 \tabularnewline
$(||\delta_{xy}||, \delta_z)$ & 78.2 \tabularnewline
$(||\delta||, \arccos \delta_z/||\delta||)$ & 78.7 \tabularnewline
\midrule
$(||\delta||)$ & 60.7 \tabularnewline
$(0)$ & 38.9 \tabularnewline
\bottomrule
\end{tabular}%
}
\vspace{1ex}
\caption{\label{tab:edgelimp}
ECC on Sydney with varied edge label definition.}
\end{table}

\section{Identity Connections} \label{sec:idconn}

The formulation of ECC in Equation~\ref{eq:C1} does not treat self-loop edges in a special way. However, the success of residual networks \cite{residuals} is a strong motivation to consider adding identity skip-connections to our model and encouraging ECC in residual learning. We thus formulate ECC-resnet as follows:

\begin{equation}
X^l(i) = \frac{1}{|N(i)|} \sum_{j\in N(i)} \Theta_{ji}^l X^{l-1}(j) + b^l + \mathrm{id}(X^{l-1}(i)) \label{eq:Crn}
\end{equation}

where $\mathrm{id}()$ is an identity mapping if $d_l=d_{l-1}$ and a linear mapping otherwise.

The results listed in Table~\ref{tab:idconntable} show that with two exceptions (NCI109 and ENZYMES) ECC does not benefit from identity connections in the specific network configurations. The trend may be different for other configurations, \eg ECC $1.5\rho$ improved from 76.9 to 79.5 mean F1 score on Sydney due to identity connections as mentioned in Section~\ref{subsec:evalpc}.

\begin{table}[bt]
\centering
\resizebox{1\linewidth}{!}{%
\begin{tabular}{cccccccc}
\toprule
 & NCI1 & NCI109 & MUTAG & ENZYMES & D{\&}D & Sydney & ModelNet10\tabularnewline
\midrule
ECC-resnet & 83.24  & \emph{81.97 } & 85.56  & \emph{51.83 } & 70.48 & 77.0 & 88.5 (89.3) \tabularnewline
ECC & 83.80  & 81.87  & 89.44  & 50.00  & 73.65  & 78.4  & 89.3 (90.0) \tabularnewline
\bottomrule
\end{tabular}%
}
\vspace{1ex}
\caption{\label{tab:idconntable}
The effect of adding identity connections (improvements in italics). Performance metrics vary and are specific to each dataset, as introduced in the main paper.}
\end{table}

\section{Vertex Degrees in Edge Labels} \label{sec:degrees}

In the task of graph classification, we used categorical labels (if present) encoded as one-hot vectors for edges in the input graph and scalars computed by Kron reduction for edges in all coarsened graphs. 

Here we investigate making the edge labels more informative by including the degrees of the pair of vertices forming an edge. The degree information is implicitly used by spectral convolution methods, as the degree information is contained in the graph Laplacian, and also appears in the explicit propagation rules \cite{kipf, dcnn}. 

Our model can be easily extended to make use of this information by simply appending it to the existing edge label vectors. We consider four variants of providing additional degree labels $L_{deg}(j)$ and  $L_{deg}(i)$ about a directed edge $(j,i)$: $L_{deg}(i) = 1/\sqrt{\mathrm{deg}(i)}$, $L_{deg}(i) = 1/\mathrm{deg}(i)$, $L_{deg}(i) = \sqrt{\mathrm{deg}(i)}$, and $L_{deg}(i) = \mathrm{deg}(i)$, where $\mathrm{deg}(i) = |N(i)|$ is the degree of vertex $i\in V$. We use these additional labels in all graph resolutions.


Table~\ref{tab:degrees} reveals that degree information can improve the results considerably, especially for datasets without given edge labels (by up to 5 percentage points for ENZYMES and up to 2.14 percentage points for D{\&}D). However, no variant of $L_{deg}(i)$ can guarantee consistent improvement over all datasets.

\begin{table}[bt]
\centering
\resizebox{1\linewidth}{!}{%
\begin{tabular}{cccccc}
\toprule
 & NCI1 & NCI109 & MUTAG & ENZYMES & D{\&}D \tabularnewline
\midrule
$L_{deg}(i) = 1/\sqrt{\mathrm{deg}(i)}$ & 82.99  & 81.94  & 87.78  & \emph{53.67 } & 73.65 \tabularnewline
$L_{deg}(i) = 1/\mathrm{deg}(i)$ & 83.60  & \emph{82.40 } & 88.89  & \emph{52.67 } & 71.77 \tabularnewline
$L_{deg}(i) = \sqrt{\mathrm{deg}(i)}$ & 83.58  & \emph{82.28 } & 86.67  & \emph{55.00 } & \emph{75.79}\tabularnewline
$L_{deg}(i) = \mathrm{deg}(i)$ & 83.16  & \emph{83.03 } & 86.67  & \emph{52.83 } & \emph{73.74 }\tabularnewline
\midrule
ECC without $L_{deg}(i)$ & 83.80  & 81.87  & 89.44  & 50.00  & 73.65 \tabularnewline
\bottomrule
\end{tabular}%
}
\vspace{1ex}
\caption{\label{tab:degrees}
The effect in mean classification accuracy of adding vertex degrees to edge labels (improvements in italics).}
\end{table}

\section{Vertex Degrees in Normalization} \label{sec:normnet}

The formulation of ECC in Equation~\ref{eq:C1} performs normalization by the neighborhood size. Here we explore learning an additional multiplicative factor, conditioned on the neighborhood size $1/|N(i)|$. To this end, we again make use of Dynamic filter networks~\cite{dfn16} and design a factor-generating network $Z^l: \mathbb{R} \mapsto \mathbb{R}$  which given vertex degree $\mathrm{deg}(i) = |N(i)|$  outputs a vertex-specific normalization factor. We formulate ECC-Z as follows:

\begin{equation}
X^l(i) = \frac{Z^l(|N(i)|;w^l)}{|N(i)|} \sum_{j\in N(i)} \Theta_{ji}^l X^{l-1}(j) + b^l \label{eq:CZ}
\end{equation}

In our experiments, the factor-generating networks $Z^l$ have configuration FC(32)-FC(1) with orthogonal weight initialization~\cite{orthoinit} and ReLUs in between. 

The results in Table~\ref{tab:normnet} show that while being helpful on some datasets (NCI109, ENZYMES, ModelNet10), ECC-Z harms the performance on the other ones. Embedding vertex information in labels instead seems to achieve higher performance (Section~\ref{sec:degrees}).

\begin{table}[bt]
\centering
\resizebox{1\linewidth}{!}{%
\begin{tabular}{cccccccc}
\toprule
 & NCI1 & NCI109 & MUTAG & ENZYMES & D{\&}D & Sydney & ModelNet10\tabularnewline
\midrule
ECC-Z & 83.48  & \emph{82.57} & 86.67  & \emph{52.50} & 72.03 & 75.5 & \emph{89.9 (90.6)} \tabularnewline 
ECC & 83.80  & 81.87  & 89.44  & 50.00  & 73.65  & 78.4  & 89.3 (90.0) \tabularnewline
\bottomrule
\end{tabular}%
}
\vspace{1ex}
\caption{\label{tab:normnet}
The effect of adding a learned normalization factor (improvements in italics). Performance metrics vary and are specific to each dataset, as introduced in the main paper.}
\end{table}

\end{document}